\documentclass[conference]{IEEEtran}
\IEEEoverridecommandlockouts
\usepackage{cite}
\usepackage{amsmath,amssymb,amsfonts}
\usepackage{algorithmic}
\usepackage{graphicx}
\usepackage{textcomp}
\usepackage{xcolor}
\usepackage{comment}
\usepackage{booktabs}
\usepackage{makecell}

\def\BibTeX{{\rm B\kern-.05em{\sc i\kern-.025em b}\kern-.08em
    T\kern-.1667em\lower.7ex\hbox{E}\kern-.125emX}}
    
\begin{document}

\title{Enhancing Clinical Information Extraction with \\Transferred Contextual Embeddings
}

\author{

\IEEEauthorblockN{
    Zimin Wan\IEEEauthorrefmark{1}, 
    Chenchen Xu\IEEEauthorrefmark{1}\IEEEauthorrefmark{2},
    Hanna Suominen\IEEEauthorrefmark{1}\IEEEauthorrefmark{2}\IEEEauthorrefmark{3}\IEEEauthorrefmark{4}
}
    
\IEEEauthorblockA{
    \IEEEauthorrefmark{1}The Australian National University (ANU) / Canberra, ACT, Australia
}
\IEEEauthorblockA{
    \IEEEauthorrefmark{2}Data61, Commonwealth Scientific and Industrial Research Organization (CSIRO) / Canberra, ACT, Australia
}
\IEEEauthorblockA{
    \IEEEauthorrefmark{3}University of Canberra / Canberra, ACT, Australia
}
\IEEEauthorblockA{
    \IEEEauthorrefmark{4}University of Turku / Turku, Finland
}
\IEEEauthorblockA{
Firstname.Lastname@anu.edu.au
}

}

\maketitle

\begin{abstract}
The Bidirectional Encoder Representations from Transformers (BERT) model has achieved the state-of-the-art performance for many natural language processing (NLP)
tasks. Yet, limited research has been contributed to studying its effectiveness when the target domain is shifted from the pre-training corpora, for example, for biomedical or clinical NLP applications. In this paper, we applied it to a widely studied a hospital information extraction (IE) task and analyzed its performance under the transfer learning setting. 
Our application became the new state-of-the-art result by a clear margin, compared with a range of existing IE models. Specifically, on this nursing handover data set, the macro-average F1 score from our model was 0.438, whilst the previous best deep learning models had 0.416. In conclusion, we showed that BERT based pre-training models can be transferred to health-related documents under mild conditions and with a proper fine-tuning process.
\end{abstract}


\section{Introduction}
\label{sec:intro}

\noindent \emph{Electronic Health} (eHealth) solutions  support patients and healthcare workers in accessing, authoring, and sharing information. 
CLEF eHealth evaluation lab has introduced tasks  as part of the \emph{Conference and Labs of the Evaluation Forum} to construct a system that converts verbal clinical handover information into electronic structured handover forms automatically \cite{suominen2015task,suominen2016task}. This processing cascades voice recording, \emph{speech recognition} (SR), \emph{information extraction} (IE), and information visualization with text editing and document sign-off by~nurses \cite{suominen2015benchmarking}. 

\begin{figure*}[t]
\centering
\includegraphics[width=\linewidth]{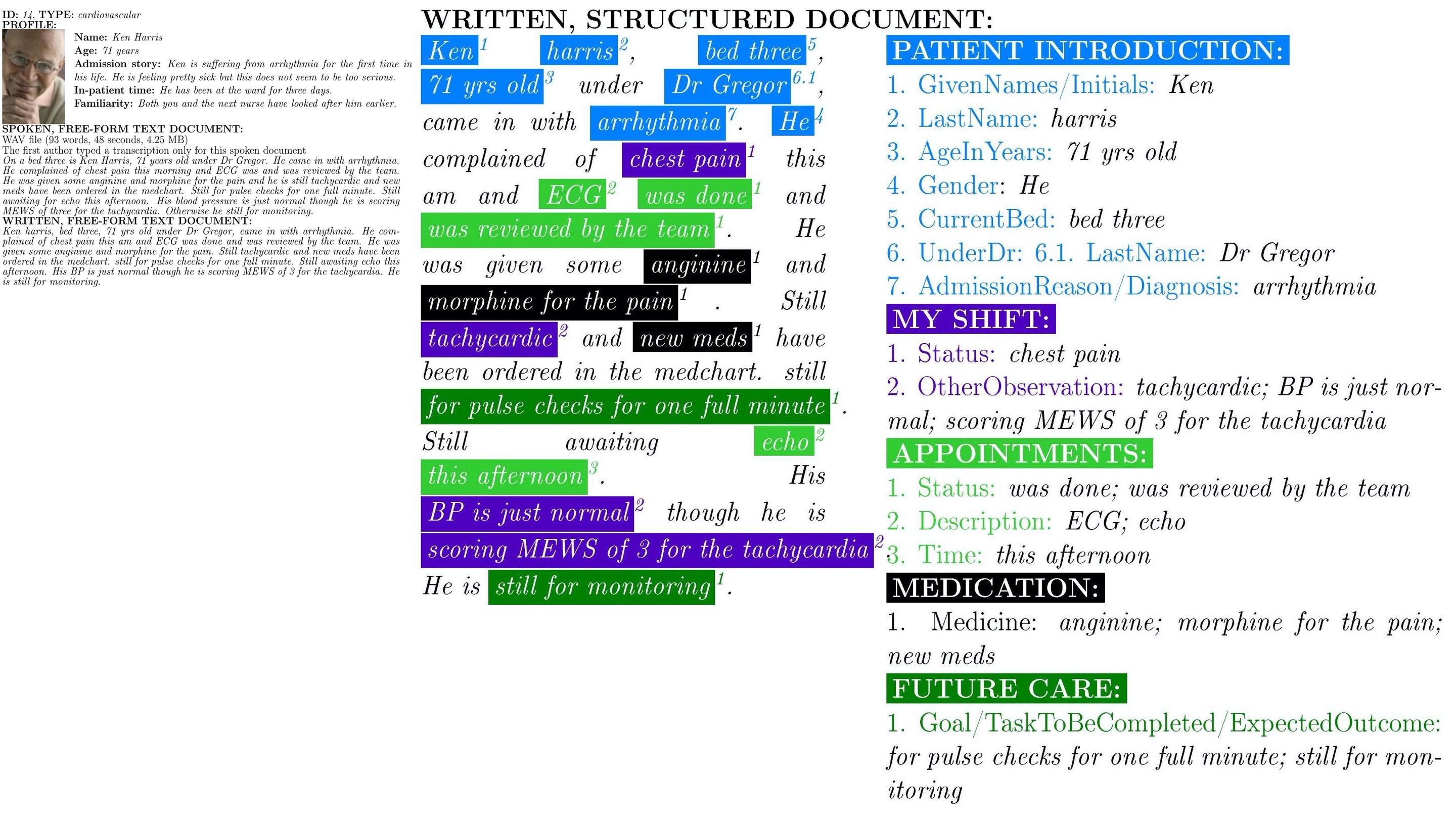}
\caption{Filling out a handover form using information extraction. Revised from \cite{suominen2015benchmarking}.}
\label{handoverfig}
\end{figure*}

In this study, we focus on this IE sub-task \cite{suominen2016task}, which is to pre-fill a shift-change nursing handover form by automatically extracting relevant text spans for each slot in the form (Figure \ref{handoverfig}). 
Although many \emph{machine learning} (ML) and \emph{natural language processing} (NLP) approaches have been applied to it, the performance is not always satisfactory \cite{suominen2015benchmarking}.
The main reasons for this gap  are the differing vocabularies and context between clinical sublanguages (e.g., nursing vs. medical wards vs. medical-ward nursing) and generic English, along with the ongoing problem of having a limited amount of healthcare-specific data available for training and evaluation; data for real-world healthcare scenarios for research and development are low-resourced \cite{VELUPILLAI201811, li2020word, zoph2016transfer}. Common methods in (deep) ML  do not achieve a satisfying  generalization to unseen data on this clinical task and consequently, the current state-of-the-art approach uses \emph{transfer learning} (TL) by adapting knowledge from general via medical to clinical English to overcome data insufficiency \cite{zhou2019adapting}. However, this remedy relies on the selection of the source domain and requires a large amount of time to pre-train the model on the source~domain.

Consequently, our aim is to analyze the adaptability to a domain shift of the timely \emph{Bidirectional Encoder Representations from Transformers} (BERT) method in the nursing handover IE task.
BERT has caused a stir in the ML-based NLP community by obtaining new state-of-the-art results in various tasks \cite{devlin2018bert}. BERT applies the bidirectional training of the Transformer model from deep learning to language modelling. 
By incorporating in-sentence word-to-word semantic relations and the cross-sentence context knowledge, it addresses the gap in linguistic understanding that limited data in typical biomedical and clinical NLP data sets are unable to provide. The subsequent fine-tuning process then provides extra domain specific knowledge in these health-related data. 

Our main contributions are twofold: 

\begin{itemize}
    \item Our application of BERT under a TL framework to the clinical IE task achieves the new state-of-the-art result. 
    \item Conducted analysis to verify our framework's capability to cope with a domain shift from the pre-training corpora to the target task with scarce data available for training and evaluation.
\end{itemize}
\section{Background}
\label{sec:background}

\noindent IE is a common task in NLP that aims to automatically extract structured information from  free-form or semi-structured text. Template filling is a subtask in IE which is to  fill out a structured form by extracting text spans. It could be seen as an ML task called sequence tagging or labelling. Its major difficulty is to build a good language model to understand language representation. ML can learn features and representations from big data,  but rich, ambiguous  human language  is a difficult data modality to understand. A word embedding is a distributed representation of words that models a given language by taking the word context into account and embeds words into vectors \cite{mikolov2013distributed}. 

A \emph{conditional random field} (CRF)  is a probabilistic sequence model, which can learn the feature representation of the words, while taking the context of words into account \cite{lafferty2001conditional}. Linear chain CRFs are popular in NLP to solve a sequence tagging tasks, and hence, their CRF++ implementation \cite{kudo2015crf++} is used as 
the benchmark method in the handover IE task with 
both unigrams and bigrams (which combine predicted label for the previous word and features of the current word) 
 and the following features~\cite{suominen2015benchmarking,suominen2016task}: the previous, current, and next location; pairwise correlations of the previous and current location and the current and next location; and the combination of all features in the current~location. 

TL is a (deep) ML method that can learn knowledge from solving one domain and apply it to a different problem. It is becoming popular as a way to use pre-trained models as a starting point towards more task-specific solutions in NLP \cite{ruder-etal-2019-transfer} and CV \cite{Li_2020_CVPR}. 
As in deep learning, many hidden layers exist between the input and output layer; 
through the data feed-forward,
the lower layers tend to extract  common and non-domain features, while the higher layers are likely to concatenate learned features and generate higher-level concepts \cite{schmidhuber2015deep, li2020tspnet}. Thus, the learned weights can be reused if the structure of the higher layers are consistent.

\begin{figure}[htbp]
\centering
\includegraphics[width=0.85\linewidth]{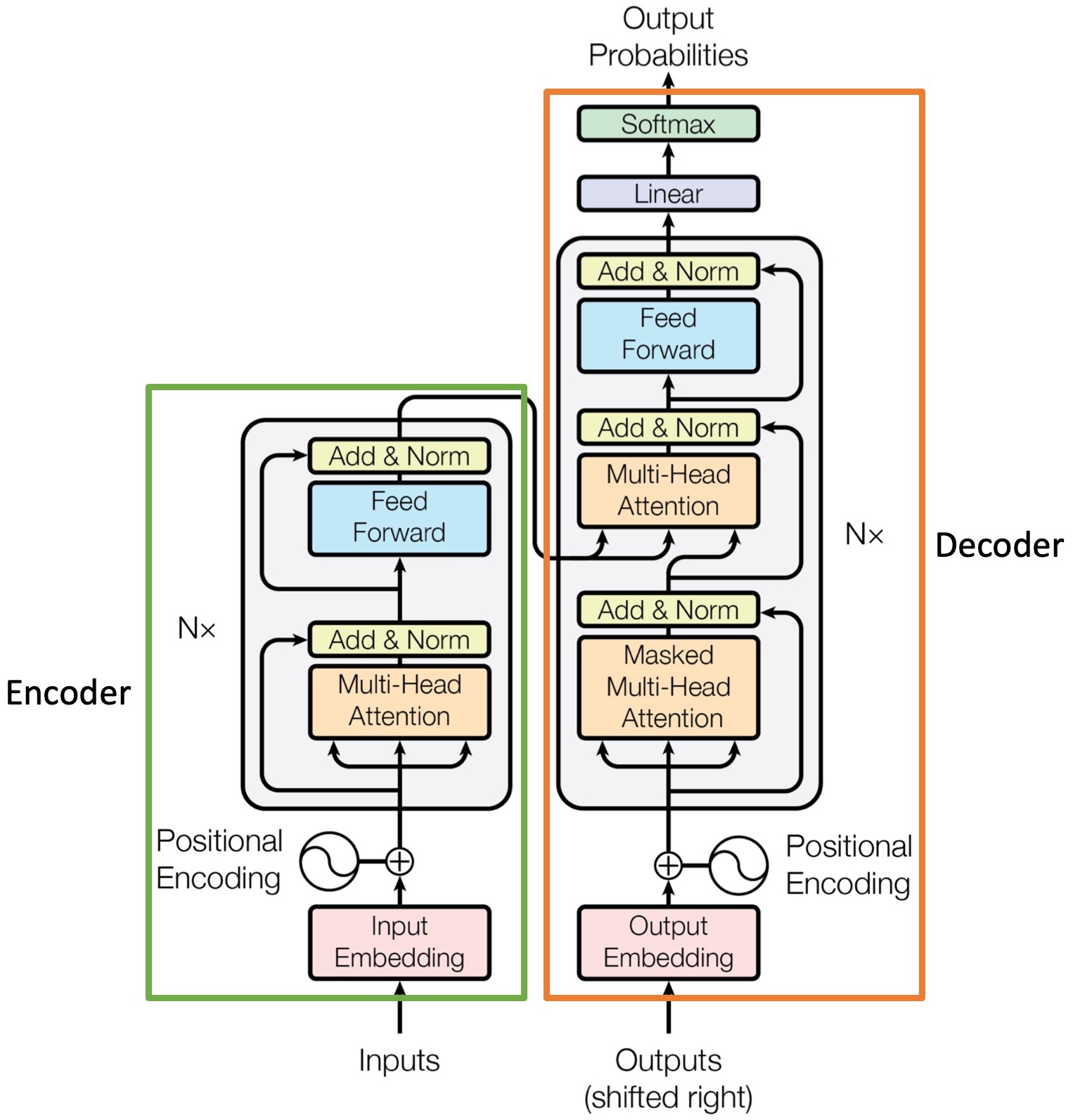}
\caption{Transformer architectures. Revised from \cite{vaswani2017attention}.}
\label{fig:transformer}
\end{figure}

In the handover IE task, TL has been applied 
to first  
train a linear chain CRF 
on the source domain as a starting point,  then, the learnt weights of the source model have been used to train a 2-layer CRF to predict a target domain label, and finally, the parameters of the target domain model have been initialized by using the product of the second layer weights and the train source model weights  \cite{zhou2019adapting}. 
Besides, the baseline model of TL uses different corpora as the source domain, and so far, the model with the best performance has used a general domain source corpus, called \emph{Bolt, Beranek and Newman} (BBN) 
\cite{weischedel2005bbn}.

\section{Materials}
\label{sec:dataset}

This study used the public NICTA Synthetic Nursing Handover Data Set, which was developed for clinical SR and IE related to nursing shift-change handover from 2012 to 2016 \cite{suominen2015benchmarking}. 
%
The data set included three mutually exclusive subsets and 301 synthetic patient records in total (training, validation, and test set with 101, 100, and 100 records, respectively) .
In this study, 
our goal was to extract relevant text spans from the written free-form text documents in order to generate the 
written structured documents.  
The structured documents included annotations of five classes,\footnote{i.e., \textit{PATIENT INTRODUCTION}, \textit{MY SHIFT}, \textit{APPOINTMENTS}, \textit{MEDICATION}, and \textit{FUTURE CARE}} 
and each class was further divided into 38 subclasses 
and an additional category of \emph{not applicable} (N.A.) for irrelevant~information.

The IE task was challenged by the nursing notes not always being consistent with formal general written English. For example, the data contained clinical jargon and (ambiguous ad-hoc) shorthand (\textit{yrs} for \textit{years} and \textit{BP} for \textit{Bachelor of Pharmacy}, \textit{bedpan}, \textit{before present}, \textit{birthplace}, or \textit{blood~pressure}). Text spans relevant to  the same subclass were also sometimes scattered across the text document. 
Finally,  
for some labels, as few as two or three instances were present in the training set.\footnote{e.g., \textit{Clinician's Last Name} and \textit{Ward}} 
In fact, not all documents included relevant information for all form categories
and 
some documents contained variants of the desired information multiple times.\footnote{e.g., for a patient whose given name was \textit{Timothy}, also his nickname of \textit{Tim} was supposed to be extracted to populate \textit{PATIENT INTRODUCTION: Given names/initials}} 

\section{Processing Methods}
\label{sec:method}

\textit{Attention} and \textit{Transformer} mechanisms are of the significant breakthroughs in language modelling, which have greatly promoted the development of NLP \cite{vaswani2017attention} and CV \cite{Li_2021_CVPR}: The former was introduced to solve the bottleneck of information pass caused by the conversion of long sequences to fixed-length vectors with \textit{recurrent neural networks} (RNNs).
The latter
uses self-attention mechanisms and completely abandons the architecture of RNNs, which can model long term dependencies among tokens in a temporal sequence more~effectively (Figure~\ref{fig:transformer}).

As Transformer contains no recurrence and convolution layers, the position information should be injected to enable the model to take advantage of the sequence order. Thus, its input is the sum of input embeddings and the positional encodings.

Transformer can model long term dependencies of a sequence and implement parallel computing to increase training speed by using a self-attention~mechanism:
In the encoder stack, each of the $N = 6$ identical layers consists of a multi-head self-attention mechanism and a position-wise fully connected feed-forward network. Besides, each sub-layer uses the residual connection followed by a normalization layer at its output.
The decoder stack is similar to the encoder stack, with $N = 6$ identical layers. It inserts a multi-head attention sub-layer over the output of the encoder. Similarly, each sub-layer uses the residual connection followed by a normalization layer at its output.

The BERT model 
is designed to pre-train deep bidirectional language representations from unlabeled text. It has become the new state-of-the-art in at least eleven NLP tasks \cite{devlin2018bert}, and hence, in this study, we have used it as the language representation model.
%
Transformer is used in BERT to learn contextual relations between words in a text sequence. Because BERT aims to generate a language representation model, it only takes use of Transformer's encoder stack. 
BERT pre-trains deep bidirectional representations from unlabeled text by jointly conditioning on both left and right context in all layers. 
%

The following two sizes of the pre-trained BERT model exist \cite{devlin2018bert}: 
\begin{enumerate}
\item \textit{BERT$_{\textrm{BASE}}$} with the \textit{number of transformer blocks} $L = 12$, \textit{hidden layer size} $H = 768
$, \textit{attention heads} $A = 12$, and \textit{number of total parameters} $P = 110$ million and 
\item \textit{BERT$_{\textrm{LARGE}}$} with $L = 24$, $H = 1,024$, $A = 16$ and $P = 340$ million. In this study, we have used the BERT$_{\textrm{BASE}}$ model.
\end{enumerate}

\begin{figure}[htpb]
\centering
\includegraphics[width=\linewidth]{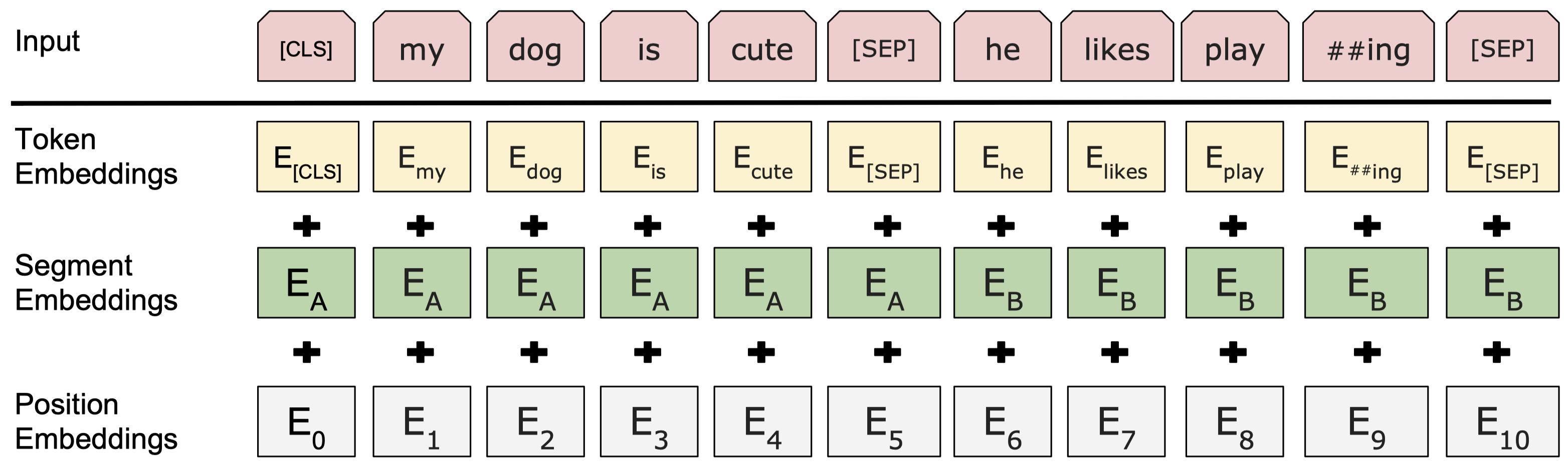}
\caption{BERT input representation} 
\label{fig:bertinput}
\end{figure}

Generally, BERT takes either one or two sentences as an input, uses \textit{WordPiece embeddings} \cite{wu2016google} to convert the sentence into a token list, and expects special tokens \textit{[CLS]} and \textit{[SEP]} 
to identify the sentences.
WordPiece embeddings  perform textual preprocessing to convert the free-form text to a list of tokens in order to
balance the flexibility of characters and  efficiency of words.
One of the main implementations of WordPiece is the \emph{Byte Pair Encoding} (BPE) double-byte encoding. The general training process of BPE is first to divide the word into characters, second, to count the number of occurrences of the character pair within the range of the word, and third, to iteratively add the most frequent character combinations to the vocabulary. Thus, BPE can split a word further so that the vocabulary will be streamlined and the meaning of each word will be clearer. 
For a given token, BERT's \textit{input representation} is the sum of the corresponding \textit{token}, \textit{segmentation}, and \textit{position embeddings} as illustrated in Figure~\ref{fig:bertinput}. 
The token embeddings are generated by WordPiece, the segment embeddings are the sentence identifiers, and the position embeddings are used to show the token position within the sequence.

\begin{figure}[htbp]
\centering
\includegraphics[width=0.85\linewidth]{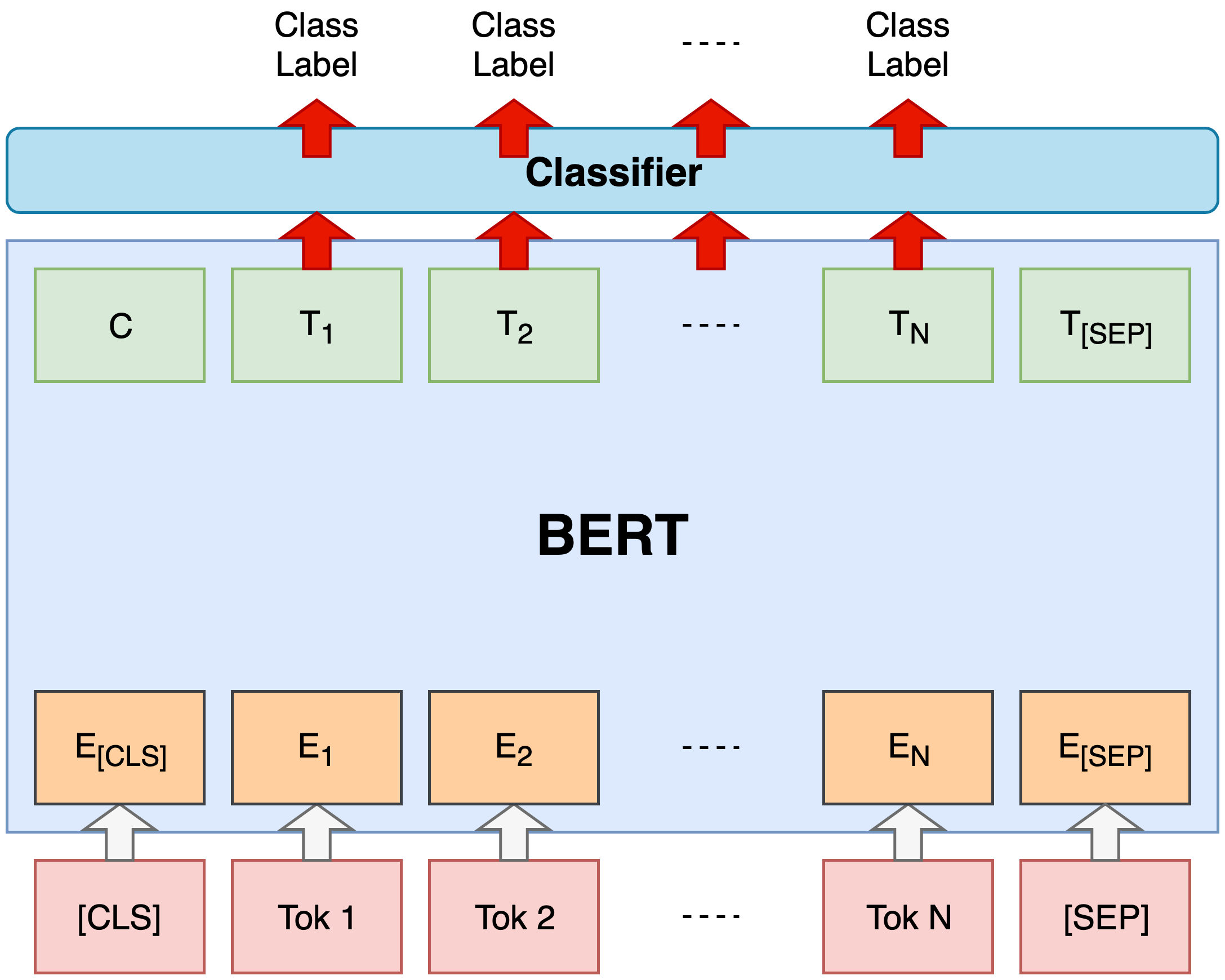}
\caption{BERT with a multi-class classifier architecture.}
\label{fig:bertclassifier}
\end{figure}

The \textit{output representation} is the word embeddings generated by the pre-trained networks. By using BERT, they will be 768-dimensional vectors, which can be then used as features in ML-based NLP models.

In IE and sequence labelling tasks, the model will receive a text sequence, and label each token to one of the various classes. A sequence model can be simply added on top of BERT by connecting BERT's hidden-states output with a token classifier (Figure~\ref{fig:bertclassifier}).
It is trainable by feeding the output representation of each token into the classification layer to predict the corresponding label. 
We add a multi-class classification layer followed by sequential cross entropy loss as the optimization target.
Fine-tuning all the parameters end-to-end to model the IE downstream task with BERT is straightforward. Most hyperparameters for the fine-tuning process are the same as in pre-training; and only learning rate, batch size, and the number of training epochs are chosen as according to the validation set \cite{devlin2018bert}.


\section{Evaluation Methods}

We have evaluated models using the Precision, Recall, and F$_1$, together with confusion matrices.
The macro-average is computed over the multiple classes
to emphasize the significance of performing well in all classes, including also those with relatively rare occurrence frequency, due to 
the risk averse nature of clinical judgement.

Because some categories in the training set do not appear in the test set and vice versa,\footnote{e.g., \textit{City} vs. \textit{Hospital} or \textit{Clinician's Title} under \textit{APPOINTMENTS}} we have calculated
the performance of proposed model on the test set based on the categories that occur in the training~set.
This follows the conventions of the earlier studies \cite{suominen2015benchmarking,suominen2016task,zhou2019adapting}.



\begin{table}[htb]
\caption{Comparison to baseline models.\\
Because of not having access
to the system outputs of the 
second best method \cite{zhou2019adapting}, assessing statistical significance was not possible.
Random and Majority refer to assigning class labels randomly and using the most common class of \textit{FUTURE: TaskToBeCompleted / ExpectedOutcome}
for every instance. Acronyms: \emph{Macro-averaged} (MAd)}
\label{tab:resultcomparison}

\centering
\begin{tabular}{lccc}
  \hline
  Method & MAd Precision & MAd Recall & MAd F$_1$ \\
  \hline
  BERT-based & 0.485 & \textbf{0.477} & \textbf{0.438} \\
  TL using BBN  & \textbf{0.498} & 0.419 & 0.416 \\
  CRF & 0.435 & 0.233 & 0.246 \\
  Random & 0.018 & 0.028 & 0.019 \\
  Majority & 0.000 & 0.029 & 0.001 \\
  \hline
\end{tabular}

\end{table}
\begin{table}[htb]
\centering
\caption{Top performing subclasses with F$_1$ score $>$ 0.8. Acronyms: \textit{PatientIntroduction }(PI)}
\label{tab:topperformance}
\begin{tabular}{ lr }
  \hline
  Subclasses & F$_1$ \\
  \hline
  PI\_CurrentRoom & 1.0000 \\
  PI\_Lastname & 0.9950 \\
  PI\_GivenNames/Initials & 0.9900 \\
  PI\_CurrentBed & 0.9645 \\
  PI\_Ageinyears & 0.9444 \\
  MyShift\_Input/Diet & 0.8618 \\
  MyShift\_ActivitiesOfDailyLiving & 0.8604 \\
  N.A. & 0.8177 \\
  PI\_UnderDr\_Lastname & 0.8127 \\
  \hline
\end{tabular}
\end{table}

\section{Results}
\label{sec:result}

\newcommand{\tabincell}[2]{\begin{tabular}{@{}#1@{}}#2\end{tabular}}

\begin{table*}[htbp]
\centering
\caption{The confusion matrix of BERT-based model on different categories.}
\label{cm}

\begin{tabular}{ p{6cm}rrrr }
  \toprule
  \textbf{CATEGORY} &
  \textbf{\tabincell{c}{Number of Words \\in Test Set}} &
  \textbf{\tabincell{c}{Number of \\True Positive}} &
  \textbf{\tabincell{c}{Number of \\False Positive}} &
  \textbf{\tabincell{c}{Number of \\False Negative}} \\
   
  \midrule[1pt]
  \textbf{A. PATIENT INTRODUCTION} & \textbf{1,\,611} & \textbf{1,\,134} & \textbf{579} & \textbf{477} \\
 
  1. Given Names/ Initials & 100 & 99 & 1 & 1 \\
   
  2. Last Name & 101 & 100 & 0 & 1 \\
   
  3. Age in Years & 281 & 280 & 32 & 1 \\
   
  4. Gender & 178 & 62 & 0 & 116 \\
   
  5. Current Room & 100 & 100 & 0 & 0 \\
   
  6. Current Bed & 198 & 190 & 6 & 8 \\
   
  7. Under Dr: Given Names/ Initials & 53 & 0 & 0 & 53 \\
   
  8. Under Dr: Last Name & 128 & 128 & 59 & 0 \\
   
  9. Admission Reason/ Diagnosis & 153 & 146 & 341 & 7 \\
   
  10. Allergy & 38 & 18 & 1 & 20 \\
   
  11. Chronic Condition & 73 & 4 & 14 & 69 \\
   
  12. Disease/ Problem History & 206 & 5 & 110 & 201 \\
   
  13. Care Plan & 2 & 2 & 15 & 0 \\
  \hline
  \textbf{B. MY SHIFT} & \textbf{1,\,033} & \textbf{717} & \textbf{606} & \textbf{316} \\
 
  14. Status & 285 & 234 & 123 & 51 \\
   
  15. Contraption & 43 & 7 & 172 & 36 \\
   
  16. Input/ Diet & 107 & 106 & 33 & 1 \\
   
  17. Output/ Diuresis/ Bowel Movement & 50 & 26 & 32 & 24 \\
   
  18. Wounds/ Skin & 25 & 20 & 34 & 5 \\
   
  19. Activities of Daily Living & 295 & 262 & 52 & 33 \\
   
  20. Risk Management & 40 & 0 & 4 & 40 \\
   
  21. Other Observation & 188 & 62 & 156 & 126 \\
  \hline
  \textbf{C. APPOINTMENTS} & \textbf{472} & \textbf{143} & \textbf{227} & \textbf{329} \\
 
  22. Status & 79 & 19 & 147 & 60 \\
   
  23. Description & 313 & 91 & 44 & 222 \\
   
  24. Clinician: Given Names/ Initials & 1 & 0 & 0 & 1 \\
   
  25. Clinician: Last name & 1 & 0 & 1 & 1 \\
   
  26. Date and Time: Day & 34 & 14 & 22 & 20 \\
   
  27. Date and Time: Time & 39 & 17 & 12 & 22 \\
   
  28. Date and Time: City & 0 & 0 & 1 & 0 \\
   
  29. Date and Time: Ward & 5 & 2 & 0 & 3 \\
  \hline
  \textbf{D. MEDICATION} & \textbf{493} & \textbf{138} & \textbf{95} & \textbf{355} \\
  30. Medicine & 296 & 120 & 65 & 176 \\
   
  31. Dosage & 52 & 8 & 18 & 44 \\
   
  32. Status & 145 & 10 & 12 & 135 \\
  \hline
  \textbf{E. FUTURE CARE} & \textbf{252} & \textbf{88} & \textbf{319} & \textbf{164} \\
 
  33. Alert/ Warning/ Abnormal Result & 28 & 8 & 8 & 20 \\
   
  34. Goal/ Task To Be Completed/ Expected~Outcome & 103 & 19 & 279 & 84 \\
   
  35. Discharge/ Transfer Plan & 121 & 61 & 32 & 60 \\
 \hline
  \textbf{F. N.A.} & \textbf{2,\,652} & \textbf{2,\,090} & \textbf{370} & \textbf{562} \\
   
  36. N.A. & 2,\,652 & 2,\,090 & 370 & 562 \\
  \bottomrule
\end{tabular}
\end{table*}
\begin{table*}[htbp]
    \caption{Performance on extracted subclasses with F$_1$ score $>$ 0.8. Acronyms: \textit{PatientIntroduction }(PI)}
    \label{tab:macro_scores}

    \begin{minipage}[t]{.5\linewidth}
      \centering

\begin{tabular}[t]{ lccc }
  \hline
    \textbf{A. PATIENT INTRODUCTION} & \textbf{0.662}  & \textbf{0.7039}  & \textbf{0.6823} \\
  \hline
1. Given Names/Initials & 0.99  & 0.99  & 0.99 \\
2. Last Name & 1.0  & 0.9901  & 0.995 \\
3. Age in Years & 0.8974  & 0.9964  & 0.9444 \\
4. Gender & 1.0  & 0.3483  & 0.5167 \\
5. Current Room & 1.0  & 1.0  & 1.0 \\
6. Current Bed & 0.9694  & 0.9596  & 0.9645 \\
7. Under Dr: Given Names/Initials & 0.0  & 0.0  & 0.0 \\
8. Under Dr: Given Last Name & 0.6845  & 1.0  & 0.8127 \\
9. Admission Reason/ Diagnosis & 0.2998  & 0.9542  & 0.4562 \\
10. Allergy & 0.9474  & 0.4737  & 0.6316 \\
11. Chronic Condition & 0.2222  & 0.0548  & 0.0879 \\
12. Disease/ Problem History & 0.0435  & 0.0243  & 0.0312 \\
13. Care Plan & 0.1174  & 1.0  & 0.2105 \\
  \hline
\textbf{B. MY SHIFT} & \textbf{0.542}  & \textbf{0.6941}  & \textbf{0.6087} \\
  \hline
14. Status & 0.6555  & 0.8211  & 0.729 \\
15. Contraption & 0.0391  & 0.1628  & 0.0631 \\
16. Input/ Diet & 0.7626  & 0.9907  & 0.8618 \\
17. Output/ Diuresis/ Bowel Movement & 0.4483  & 0.52  & 0.4815 \\
18. Wounds/ Skin & 0.3704  & 0.8  & 0.5063 \\
19. Activities of Daily Living & 0.8344  & 0.8881  & 0.8604 \\
20. Risk Management & 0.0  & 0.0  & 0.0 \\
21. Other Observation & 0.2844  & 0.3298  & 0.3054 \\
  \hline
\end{tabular}

    \end{minipage}%
    \begin{minipage}[t]{.5\linewidth}
      \centering
        
\begin{tabular}[t]{ lccc }
  \hline
\textbf{C. APPOINTMENTS} & \textbf{0.3865}  & \textbf{0.303}  & \textbf{0.3397} \\
  \hline
22. Status & 0.1145  & 0.2405  & 0.1551 \\
23. Description & 0.6741  & 0.2907  & 0.4062 \\
24. Clinician: Given Names/ Initials & 0.0  & 0.0  & 0.0 \\
25. Clinician: Last Name & 0.0  & 0.0  & 0.0 \\
26. Day & 0.3889  & 0.4118  & 0.4 \\
27. Time & 0.5862  & 0.4359  & 0.5 \\
28. City & 0.0  & 0.0  & 0.0 \\
29. Ward & 1.0  & 0.4  & 0.5714 \\
  \hline
\textbf{D. MEDICATION} & \textbf{0.5923}  & \textbf{0.2799}  & \textbf{0.3802} \\
  \hline
30. Medicine & 0.6486  & 0.4054  & 0.499 \\
31. Dosage & 0.3077  & 0.1538  & 0.2051 \\
32. Status & 0.4545  & 0.06897  & 0.1198 \\
  \hline
\textbf{E. FUTURE CARE} & \textbf{0.2162}  & \textbf{0.3492}  & \textbf{0.2671} \\
  \hline
33. Alert/ Warning/ Abnormal Result & 0.6586  & 0.4054  & 0.499 \\
34. Goal/ Task To Be &  0.3077  & 0.1538  & 0.2051 \\ 
Completed/ Expected Outcome & & & \\
35. Discharge/ Transfer Plan & 0.4545  & 0.06897  & 0.1198 \\
  \hline
\textbf{F. N.A.} & \textbf{0.8496}  & \textbf{0.7881}  & \textbf{0.8177} \\
  \hline
36. N.A. & 0.8496  & 0.7881  & 0.8177 \\
  \hline
\textbf{Total} & \textbf{0.4850}  & \textbf{0.4770}  & \textbf{0.4380} \\
  
  \hline
\end{tabular}
   
    \end{minipage} 

\end{table*}


Overall, our fined-tuned model 
had the macro-average precision, recall, and 
F$_1$ score of
48.5\%, 47.7\%, and~43.8\%, respectively (Table \ref{tab:resultcomparison}). 
It excelled in
nine subclasses out of the 36 sub-classes with F$_1 \geq 0.8$ (Table \ref{tab:topperformance}).
Its performance was good for the main categories of PATIENT INTRODUCTION (F$_1 = 68.2\%$) and MY SHIFT (F$_1 = 60.9\%$). Detailed performance result per category is shown in Figure~\ref{tab:macro_scores} and our proposed model is able to achieve promising result even in minor categories which only contain minimal number of examples in the training set. Table~\ref{cm} demonstrates the confusion matrix result from our model.

When benchmarked against the previous works, our fine-tuned resulting model outperformed the previous state-of-the-art TL method using BNN as a source domain and CRF based method. It outperformed the second best method by 2.2\% for Macro $F_1$ score, with both promising precision and recall rate (Table \ref{tab:resultcomparison}). 


  



\section{Conclusion}
\label{sec:conc}

The BERT-based model became the new state-of-the-art in the clinical handover task.
It performed well in IE of names, numbers, and some behaviour of the patient with their own specialized, distinguishable vocabulary\footnote{e.g., diet and daily activities} but failed with complex descriptive content.\footnote{e.g., the disease and problem history and chronic condition, which are rare (common) in the generic English (clinical), were difficult to classify, because BERT's pre-training data.}
TL could partially overcome the training problem of CRF with small data by learning features and reusing the weights pre-trained from other domains (i.e., general English) to adapt for the current task \cite{zhou2019adapting}.

Our fine-tuned model outperforming the prior state-of-the-art illustrates the usefulness of TL of language representations for biomedical NLP applications. The biomedical corpora embeds unique word distributions and different semantics that are mostly not visible in the general corpora used to train initial BERT model. We show that the domain shift can be properly mitigated by a fine-tuning process, giving superior performance without any handcrafting. 

However, we believe there are still spaces left to explore on the model architecture to reduce the potential noises introduced by the pre-trained language model. For example, a CRF could be applied at the top layer of BERT instead of a fully-connected layer. This would make it possible to plug in other available clinical features, while in this study, we only used text data. In the meantime, we also propose to validate the capability of TL such models to more diverse applications in bioinformatics, medicine, health sciences, healthcare, and other health-related fields.

%





\newpage



\bibliographystyle{IEEEtran}
\bibliography{conference_101719}

\begin{thebibliography}{10}
\providecommand{\url}[1]{#1}
\csname url@samestyle\endcsname
\providecommand{\newblock}{\relax}
\providecommand{\bibinfo}[2]{#2}
\providecommand{\BIBentrySTDinterwordspacing}{\spaceskip=0pt\relax}
\providecommand{\BIBentryALTinterwordstretchfactor}{4}
\providecommand{\BIBentryALTinterwordspacing}{\spaceskip=\fontdimen2\font plus
\BIBentryALTinterwordstretchfactor\fontdimen3\font minus
  \fontdimen4\font\relax}
\providecommand{\BIBforeignlanguage}[2]{{%
\expandafter\ifx\csname l@#1\endcsname\relax
\typeout{** WARNING: IEEEtran.bst: No hyphenation pattern has been}%
\typeout{** loaded for the language `#1'. Using the pattern for}%
\typeout{** the default language instead.}%
\else
\language=\csname l@#1\endcsname
\fi
#2}}
\providecommand{\BIBdecl}{\relax}
\BIBdecl

\bibitem{suominen2015task}
H.~Suominen, L.~Hanlen, L.~Goeuriot, L.~Kelly, and G.~J. Jones, ``Task 1a of
  the clef ehealth evaluation lab 2015: Clinical speech recognition,'' in
  \emph{CLEF 2015 Working notes.}, 2015.

\bibitem{suominen2016task}
H.~Suominen, L.~Zhou, L.~Goeuriot, and L.~Kelly, ``Task 1 of the {CLEF eHealth
  Evaluation Lab} 2016: Handover information extraction.'' in \emph{Working
  Notes of Conference and Labs of the Evaluation Forum (CLEF)}, ser. CLEF2016
  Working Notes, K.~Balog, L.~Cappellato, N.~Ferro, and C.~Macdonald, Eds.,
  vol. 1609.\hskip 1em plus 0.5em minus 0.4em\relax Online: CEUR Workshop
  Proceedings, 2016, pp. 1--14.

\bibitem{suominen2015benchmarking}
H.~Suominen, L.~Zhou, L.~Hanlen, and G.~Ferraro, ``Benchmarking clinical speech
  recognition and information extraction: new data, methods, and evaluations,''
  \emph{JMIR medical informatics}, vol.~3, no.~2, p. e19, 2015.

\bibitem{VELUPILLAI201811}
S.~Velupillai, H.~Suominen, M.~Liakata, A.~Roberts, A.~D. Shah, K.~Morley,
  D.~Osborn, J.~Hayes, R.~Stewart, J.~Downs, W.~Chapman, and R.~Dutta, ``Using
  clinical natural language processing for health outcomes research: Overview
  and actionable suggestions for future advances,'' \emph{Journal of Biomedical
  Informatics}, vol.~88, pp. 11--19, 2018.

\bibitem{li2020word}
D.~Li, C.~Rodriguez, X.~Yu, and H.~Li, ``Word-level deep sign language
  recognition from video: A new large-scale dataset and methods comparison,''
  in \emph{The IEEE Winter Conference on Applications of Computer Vision},
  2020, pp. 1459--1469.

\bibitem{zoph2016transfer}
B.~Zoph, D.~Yuret, J.~May, and K.~Knight, ``Transfer learning for low-resource
  neural machine translation,'' in \emph{Proceedings of the 2016 Conference on
  Empirical Methods in Natural Language Processing}, 2016, pp. 1568--1575.

\bibitem{zhou2019adapting}
L.~Zhou, H.~Suominen, and T.~Gedeon, ``Adapting state-of-the-art deep language
  models to clinical information extraction systems: Potentials, challenges,
  and solutions,'' \emph{JMIR Medical Informatics}, vol.~7, no.~2, p. e11499,
  2019.

\bibitem{devlin2018bert}
J.~Devlin, M.-W. Chang, K.~Lee, and K.~Toutanova, ``Bert: Pre-training of deep
  bidirectional transformers for language understanding,'' in \emph{Proceedings
  of the 2019 Conference of the North American Chapter of the Association for
  Computational Linguistics: Human Language Technologies, Volume 1 (Long and
  Short Papers)}, 2019, pp. 4171--4186.

\bibitem{mikolov2013distributed}
T.~Mikolov, I.~Sutskever, K.~Chen, G.~S. Corrado, and J.~Dean, ``Distributed
  representations of words and phrases and their compositionality,'' in
  \emph{Advances in Neural Information Processing Systems}, 2013, pp.
  3111--3119.

\bibitem{lafferty2001conditional}
J.~D. Lafferty, A.~McCallum, and F.~C.~N. Pereira, ``Conditional random fields:
  Probabilistic models for segmenting and labeling sequence data,'' in
  \emph{Proceedings of the Eighteenth International Conference on Machine
  Learning}, ser. ICML '01.\hskip 1em plus 0.5em minus 0.4em\relax San
  Francisco, CA, USA: Morgan Kaufmann Publishers Inc., 2001, pp. 282--289.

\bibitem{kudo2015crf++}
T.~Kudo, ``{CRF++: Yet another CRF toolkit (2005)},'' \emph{Available under
  LGPL from the following URL: \url{http://crfpp. sourceforge.net}}, 2015.

\bibitem{ruder-etal-2019-transfer}
\BIBentryALTinterwordspacing
S.~Ruder, M.~E. Peters, S.~Swayamdipta, and T.~Wolf, ``Transfer learning in
  natural language processing,'' in \emph{Proceedings of the 2019 Conference of
  the North {A}merican Chapter of the Association for Computational
  Linguistics: Tutorials}.\hskip 1em plus 0.5em minus 0.4em\relax Minneapolis,
  Minnesota: Association for Computational Linguistics, Jun. 2019, pp. 15--18.
  [Online]. Available: \url{https://www.aclweb.org/anthology/N19-5004}
\BIBentrySTDinterwordspacing

\bibitem{Li_2020_CVPR}
D.~Li, X.~Yu, C.~Xu, L.~Petersson, and H.~Li, ``Transferring cross-domain
  knowledge for video sign language recognition,'' in \emph{Proceedings of the
  IEEE/CVF Conference on Computer Vision and Pattern Recognition (CVPR)}, June
  2020.

\bibitem{schmidhuber2015deep}
J.~Schmidhuber, ``Deep learning in neural networks: An overview,'' \emph{Neural
  Networks}, vol.~61, pp. 85--117, 2015.

\bibitem{li2020tspnet}
D.~Li, C.~Xu, X.~Yu, K.~Zhang, B.~Swift, H.~Suominen, and H.~Li, ``Tspnet:
  Hierarchical feature learning via temporal semantic pyramid for sign language
  translation,'' in \emph{Advances in Neural Information Processing Systems},
  vol.~33, 2020.

\bibitem{vaswani2017attention}
A.~Vaswani, N.~Shazeer, N.~Parmar, J.~Uszkoreit, L.~Jones, A.~N. Gomez,
  {\L}.~Kaiser, and I.~Polosukhin, ``Attention is all you need,'' in
  \emph{Advances in Neural Information Processing Systems}, 2017, pp.
  5998--6008.

\bibitem{weischedel2005bbn}
\BIBentryALTinterwordspacing
R.~Weischedel and A.~Brunstein, ``{BBN} pronoun coreference and entity type
  corpus,'' \emph{Linguistic Data Consortium, Philadelphia}, vol. 112, 2005.
  [Online]. Available: \url{https://catalog.ldc.upenn.edu/LDC2005T33}
\BIBentrySTDinterwordspacing

\bibitem{Li_2021_CVPR}
D.~Li, C.~Xu, K.~Zhang, X.~Yu, Y.~Zhong, W.~Ren, H.~Suominen, and H.~Li,
  ``Arvo: Learning all-range volumetric correspondence for video deblurring,''
  in \emph{Proceedings of the IEEE/CVF Conference on Computer Vision and
  Pattern Recognition (CVPR)}, June 2021, pp. 7721--7731.

\bibitem{wu2016google}
Y.~Wu, M.~Schuster, Z.~Chen, Q.~V. Le, M.~Norouzi, W.~Macherey, M.~Krikun,
  Y.~Cao, Q.~Gao, K.~Macherey \emph{et~al.}, ``Google's neural machine
  translation system: Bridging the gap between human and machine translation,''
  \emph{arXiv preprint arXiv:1609.08144}, 2016.

\end{thebibliography}

\end{document}